\documentclass[11pt]{article}

\usepackage{acl}

\usepackage{times}
\usepackage{latexsym}
\usepackage[T1]{fontenc}
\usepackage[utf8]{inputenc}
\usepackage{microtype}
\usepackage{inconsolata}
\usepackage{graphicx}
\usepackage{booktabs}
\usepackage{amsmath}
\usepackage{amssymb}
\usepackage{multirow}
\usepackage{array}
\usepackage{xcolor}
\usepackage{url}
\usepackage{float}
\usepackage{algorithm}
\usepackage{algpseudocode}
\usepackage{tabularx}
\usepackage{pifont}
\usepackage{ragged2e}
\usepackage{array}
\newcommand{\sys}{\mbox{\textsc{RW-TTT}}}

\title{\sys{}: Batched Serving for Request-Owned Test-Time Training State}

\author{
\bfseries Jian Yang\textsuperscript{1} \quad Zhizhuo Kou\textsuperscript{1} \quad
Hao Zhang\textsuperscript{2} \quad Han Chen\textsuperscript{3} \\
\bfseries Yao Tian\textsuperscript{1}\NoHyper\thanks{Corresponding authors.}\endNoHyper \quad
Sirui Han\textsuperscript{1}\NoHyper\footnotemark[1]\endNoHyper \quad Yike Guo\textsuperscript{1} \\[2pt]
\normalfont\textsuperscript{1}The Hong Kong University of Science and Technology \\
\normalfont\textsuperscript{2}The Chinese University of Hong Kong \quad
\textsuperscript{3}National University of Singapore \\[2pt]
\normalfont\small\texttt{jyang827@connect.ust.hk, ytianbc@cse.ust.hk,
siruihan@ust.hk}
}

\begin{document}
\maketitle

\begin{abstract}
Test-time training (TTT) adapts an LLM during generation by reading and updating
request-owned state, such as fast weights, low-rank deltas, or streaming learner
state. This breaks batched LLM serving, which assumes shared static weights:
serial execution is correct but slow, while naive batching can corrupt request
state. We formulate this problem as \emph{read-write TTT serving} and present
\sys{}, which tags each decode step with its owner, version, and READ/WRITE
effect, batches only compatible phases, and commits updates only to the owner.
On one GPU with eight fast-weight In-Place-TTT streams, \sys{} reaches 274.61
aggregate tok/s, 9.31$\times$ over sequential serving and 3.44$\times$ over memory-matched independent replicas. It preserves behavior on
RULER, a long-context benchmark, and passes owner/version checks.
\end{abstract}

\section{Introduction}

Modern LLM serving batches requests over shared, read-only weights.
vLLM manages dynamic activation-side state---paged KV cache---but every request
still reads the same model parameters
\citep{yu2022orca,kwon2023vllm,agrawal2023sarathi,agrawal2024sarathiserve,zhong2024distserve,zheng2024sglang}.
Adapter-serving systems such as S-LoRA and Punica add request-specific
\emph{model}-side state, yet those adapters are pre-existing and read-only
during generation \citep{sheng2023slora,chen2023punica}.
Test-time training (TTT) breaks this contract
\citep{sun2020testtime,sun2024tttlayers,feng2026inplacettt}.
A TTT request may carry mutable model-side adaptation state that is both read
and \emph{updated} during decoding. In In-Place TTT, this state is a set of
request-specific fast weights in the MLP down-projection
\citep{feng2026inplacettt}. In LoRA-style test-time learning, the mutable state
can be represented as low-rank parameter deltas rather than full-model updates
\citep{hu2025test}. In end-to-end TTT for long context, the model continues
learning from the incoming context, yielding streaming learner state that changes
as the context is processed \citep{tandon2025end}. This state is request-owned: it must not leak to other requests, and its updates
must follow the backend's prescribed state transition. A serial TTT path can keep
this transition implicit by processing one stream at a time, but batched serving
must make it explicit. In particular, if a write fails, falls back to a sequential
path, or is rolled back after speculative execution, the runtime must know which
state version remains valid. We call the resulting tension \emph{read-write TTT
serving}: READ steps apply request-owned state without mutation, while WRITE
steps update the state and commit a new version.

We propose \sys{}, a serving system for this setting. \sys{} tags each serving
step with its owner and READ/WRITE effect, groups only compatible phases into
legal batches, executes read and write operators separately, and commits updates
exclusively to the owning requests. To support rollback, failed-write recovery,
fallback paths, and speculative branches, \sys{} maintains explicit state
versions: READ steps preserve version \(v\), while successful WRITE steps commit
version \(v+1\). The same contract covers the mutable-state forms above---fast
weights, low-rank deltas, and streaming learner state when the state is
independently versioned and owner-local; immediate cross-request shared writes
require additional coordination. The backend supplies the
payload and update rule; \sys{} supplies ownership, versioning, rollback-safe
commit, and batch planning.
On a long-context serving workload with a fast-weight In-Place-TTT checkpoint, \sys{} reaches 274.61 aggregate tok/s, a 9.31$\times$ gain over the sequential TTT path and a 3.44$\times$ gain over memory-matched independent replicas. RULER\citep{hsieh2024ruler} long-context evaluations show that \sys{} preserves the adapted checkpoint's task behavior.

The paper makes three contributions.
\begin{enumerate}
    \item \textbf{A read-write TTT serving contract.} We formulate the serving
    problem for online-adaptive TTT checkpoints and define its core primitives:
    request ownership, state versions, READ/WRITE effects, and
    owner-preserving commit (\S\ref{sec:serving-model}).
    \item \textbf{An owner/version-aware mutable-state batching planner.} We
    design a planner that separates READ and WRITE phases, groups only
    compatible next steps, and commits updated versions exclusively to correct
    request owners (\S\ref{sec:runtime}).
    \item \textbf{End-to-end serving evidence with behavior and correctness
    preservation.} On the evaluated fast-weight TTT checkpoint, \sys{}
    recovers 9.31$\times$ aggregate throughput over the sequential path while
    preserving RULER long-context evaluation (32K/64K) scores and passing owner/version correctness
    gates. (\S\ref{sec:experiments}).
\end{enumerate}

\begin{table*}[t]
\centering
\small
\setlength{\tabcolsep}{6pt}
\renewcommand{\arraystretch}{1.12}
\newcolumntype{Y}{>{\raggedright\arraybackslash}X}
\begin{tabularx}{\textwidth}{@{}l l c c Y@{}}
\toprule
System / abstraction
& Native request state
& Online WRITE
& Versioned commit
& Native batching key \\
\midrule
vLLM / PagedAttention
& KV cache
& --
& --
& token position / KV page \\

SGLang
& KV / prefix cache
& --
& --
& prefix or program state \\

Punica / S-LoRA
& pre-existing adapter
& --
& --
& adapter id / rank \\

Sequential TTT path
& mutable fast state
& \checkmark
& implicit / serial
& single stream \\

\sys{}
& request-owned TTTState
& \checkmark
& \checkmark
& R/W phase + version + state \\
\bottomrule
\end{tabularx}
\caption{
Serving-contract comparison. Existing systems batch read-only request state:
KV/prefix caches or pre-existing adapters. Read-write TTT instead batches
request-owned model state that mutates during decoding, requiring explicit
ownership, versioned commit, and READ/WRITE-aware batching.
}
\label{tab:serving_gap_matrix}
\end{table*}

\section{Background}

\subsection{Serving with Mutable Model State}

Autoregressive serving has two phases: \emph{prefill}, which builds the KV cache
from the prompt, and \emph{decode}, which generates tokens while extending that
cache. Existing serving systems largely rely on an assumption of immutable
model-side state.
\begin{itemize}
    \item \textbf{Static-weight serving.} Model parameters are shared across
    requests and are not modified during generation. This assumption underlies
    continuous batching in OrCA~\citep{yu2022orca}, paged KV-cache management in
    vLLM~\citep{kwon2023vllm}, and structured-generation scheduling in
    SGLang~\citep{zheng2024sglang}.
    \item \textbf{Adapter serving.} S-LoRA~\citep{sheng2023slora} and
    Punica~\citep{chen2023punica} batch requests with different
    adapters~\citep{hu2022lora,huggingfacepeft,li2025aira,li2025efficient}, but each adapter is fixed
    during generation.
\end{itemize}

Test-time training (TTT) invalidates this assumption. TTT methods adapt model
computation using inference-time information
~\citep{sun2020testtime,sun2024tttlayers,feng2026inplacettt}. From a serving
perspective, the defining property is \emph{request-owned mutable model-side
state}: each request may carry fast weights, low-rank deltas, or tail buffers
that evolve with its local context. Serving such requests requires two logical
operations:
\begin{itemize}
    \item \textbf{READ}, which applies the request's current state version during
    token computation; and
    \item \textbf{WRITE}, which updates the owning request's state, typically at
    chunk boundaries such as every $C_{\mathrm{ttt}}$ generated tokens.
\end{itemize}
Because requests arrive independently and advance at different rates, one request may be updating its state while others are still reading earlier state versions. Thus, READ and WRITE operations are not globally synchronized across requests.

\subsection{Serving Gap}

Table~\ref{tab:serving_gap_matrix} summarizes the gap. Existing runtimes manage
either request-local activation state, as in KV-cache systems, or immutable
request-specific model-side state, as in adapter-serving systems. TTT instead
requires \emph{owned, versioned, mutable} model-side state. Compatible READs can
be batched because they only observe existing versions. WRITEs, however, mutate
the owning request's state and must not become visible to incompatible READs.
A TTT serving runtime must therefore track whose state a WRITE updates, which
state version each READ observes, and when an updated version becomes visible.

\begin{table}[t]
\centering
\small
\resizebox{\columnwidth}{!}{\begin{tabular}{lcc}
\toprule
{Execution strategy} &{Streams/processes} & {Throughput} \\
\midrule
Sequential serving & 8 streams & 29.51 tok/s \\
Independent replicas & 3 batch-1 processes & 79.78 tok/s \\
\bottomrule
\end{tabular}}
\caption{Throughput of TTT serving baselines on the In-Place-TTT checkpoint~\citep{feng2026inplacettt}. Sequential execution preserves READ/WRITE semantics but limits batching, while independent replicas improve isolation but cannot share compute across compatible READs.}
\label{tab:ttt_baseline_throughput}
\end{table}

A straightforward way to support TTT is to serialize or isolate requests so that
READ and WRITE operations cannot interfere. As shown in
Table~\ref{tab:ttt_baseline_throughput}, these baselines preserve correctness but
leave substantial batching opportunities unused. Sequential serving reaches only
29.51\,tok/s with eight streams. Independent replicas improve throughput to
79.78\,tok/s using three batch-1 processes under $\sim$33\,GiB, but still treat
requests as isolated executions and therefore cannot share compute across
compatible READ phases. 

This leaves the central runtime question: \emph{how can a serving system batch
requests with mutable model-side state while preserving ownership and version
semantics?}

\section{Serving Model}
\label{sec:serving-model}

Read-write TTT serving is the problem of batching generation requests whose
model-side adaptation state is request-owned and mutable. Given active
requests $\mathcal{R}_t$, the serving system must decide which next
transitions to issue together, and when to issue them, while preserving each
request's owner, state version, read/write effect, and sequential TTT
behavior. The goal is to recover legal
batching capacity under the semantic constraints introduced by online
mutation.

A request $r$ contains ordinary inference state and mutable TTT state:
\begin{equation}
q_r = (x_r,\; k_r,\; p_r,\; s_r^v,\; m_r),
\end{equation}
where $x_r$ is the current token or hidden input, $k_r$ is KV/cache state,
$p_r$ is decode position, $s_r^v$ is the versioned request-owned TTT state,
and $m_r$ stores serving-visible metadata such as tail length and update
readiness. Base model parameters remain shared across requests; the mutable
state $s_r^v$ does not.

\paragraph{Ownership.}
Each mutable TTT state has exactly one owner. A read transition may consume
the owner's current state without changing its version:
\[
\textsc{Read}_\theta(r,x_r,s_r^v)\rightarrow (y_r,s_r^v).
\]
A write transition consumes backend update evidence $u_r$, produces a new
state version, and must commit that version back to the same owner:
\[
\textsc{Write}_\theta(r,x_r,u_r,s_r^v)\rightarrow (y_r,s_r^{v+1}).
\]
This ownership rule distinguishes TTT state from shared model weights and from
activation-only KV cache. A serving schedule may reorder independent owners
for batching, but it may not redirect, merge, or alias their mutable states.

\paragraph{Legal batching.}
Read and write transitions are never mixed in the same operator group.
Generally, a group is legal only when its requests share the same state effect,
backend state type, and operator shape class, and when each request is at the
expected committed version. The backend state type separates fast-weight TTT,
LoRA-style delta TTT, and streaming learner state. The operator shape class
records the kernel-relevant dimensions for that backend, such as dtype, layer
set, hidden dimensions, chunk size, rank, or active-row count. These keys make
the batching rule explicit: activation compatibility is not enough
when requests also read and write private model-side state.

Formally, let the next transition exposed by request $r_i$ be
\begin{equation}
e_i=(r_i,\tau_i,\sigma_i,\rho_i,v_i),
\label{eq:event}
\end{equation}
where $\tau_i$ is backend state type, $\sigma_i$ is operator shape class,
$\rho_i\in\{\textsc{Read}, \textsc{Write}\}$ is the state effect, and $v_i$
is the expected state version. At decode step $t$, the planner observes
candidate transitions
$\mathcal{E}_t=\{e_i\}$ and emits legal groups
$\mathcal{G}_t=\{G_k\}$. A group $G$ is legal only if
\begin{equation}
\begin{aligned}
\forall i,j\in G:\;&
(\tau_i,\sigma_i,\rho_i)
  =(\tau_j,\sigma_j,\rho_j),\\
&v_i=V(r_i),
\end{aligned}
\label{eq:compatibility}
\end{equation}
where $V(r_i)$ is the serving system's current committed version for owner
$r_i$. The group also carries an injective owner map: no two batch elements
may commit to the same request-owned state slot in the same transition. Thus a
legal read-write TTT batch is not merely an activation batch; it is a group of
state-effect-compatible transitions with valid owner-version bindings.

\paragraph{Bounded waiting.}
The planner may delay a ready transition to improve compatibility, but only
within a bounded wait budget. If transition $e$ for request $r$ becomes ready
at decode step $t_{\mathrm{ready}}(r,e)$ and is issued at
$t_{\mathrm{issue}}(r,e)$, then
\begin{equation}
0 \le t_{\mathrm{issue}}(r,e)-t_{\mathrm{ready}}(r,e)\le w .
\label{eq:wait-budget}
\end{equation}
Thus $w=0$ is greedy batching, and finite $w$ bounds starvation in decode
steps while permitting short phase-alignment waits. Waiting may change which
compatible requests are issued together; it may not change an individual
request's read/write order or update rule.

\paragraph{Correctness contract.}
The target behavior is the sequential TTT execution using the same
backend and update rule. An execution schedule is behavior-preserving if it
preserves request-output mapping, read-step immutability, write order and
count, version progression, owner-local commit, and tail/cache consistency for
every request. In experiments, we validate this semantic contract through
owner/version/tail invariants and by checking that per-token NLL and task
scores match the sequential TTT reference within tolerance. This
is the serving contract: \sys{} changes the execution plan, not the adapted
TTT semantics.

\section{\sys{} Serving System}
\label{sec:runtime}

\begin{figure}[t]
\centering
\includegraphics[width=0.95\linewidth]{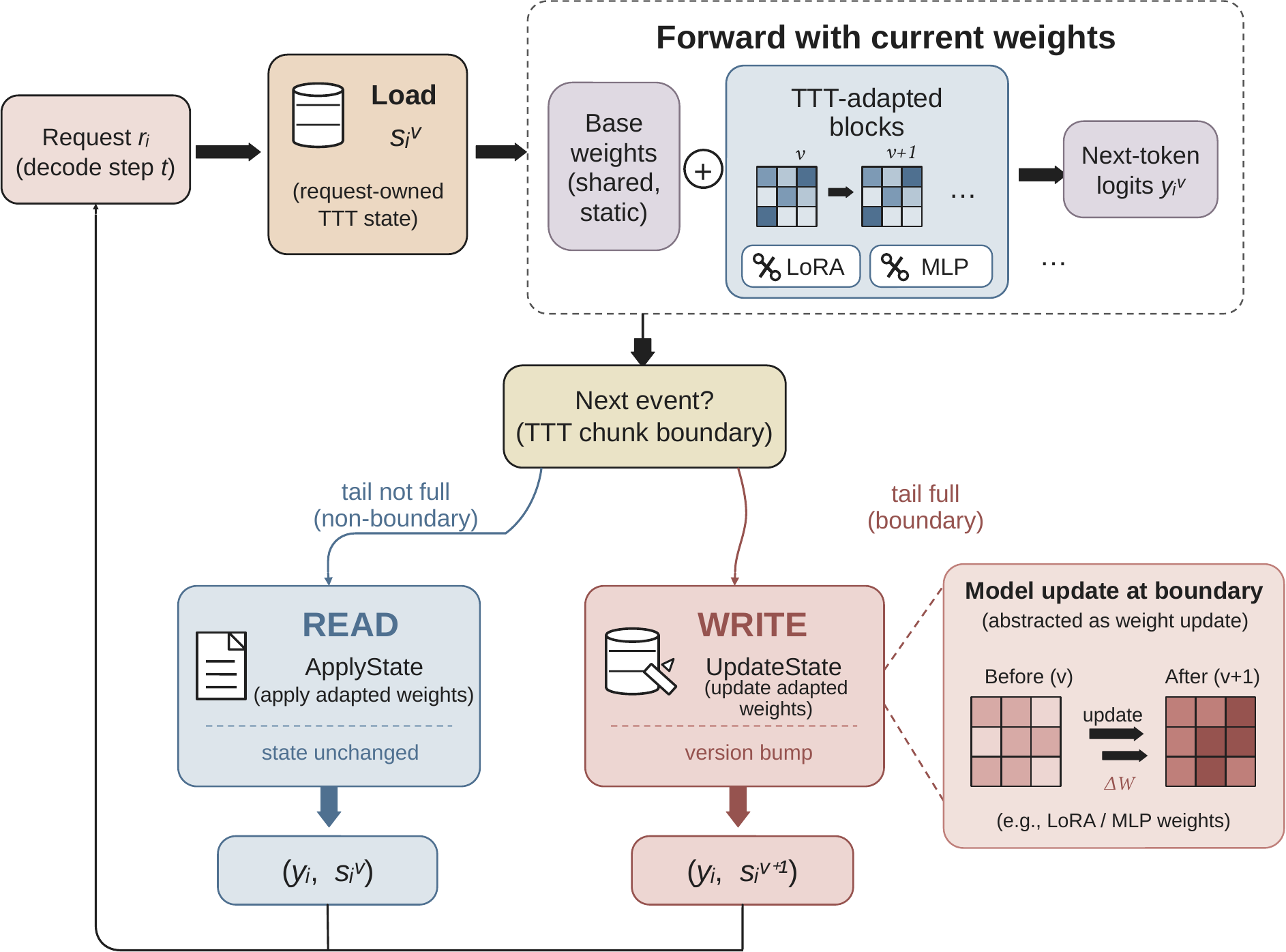}
\caption{Per-request READ/WRITE semantics. Each decode step evaluates
shared base model with request-owned TTT state and exposes either
version-preserving READ event or version-creating WRITE event.}
\label{fig:request_flow}
\end{figure}

\begin{figure*}[t]
\centering
\includegraphics[width=0.85\textwidth]{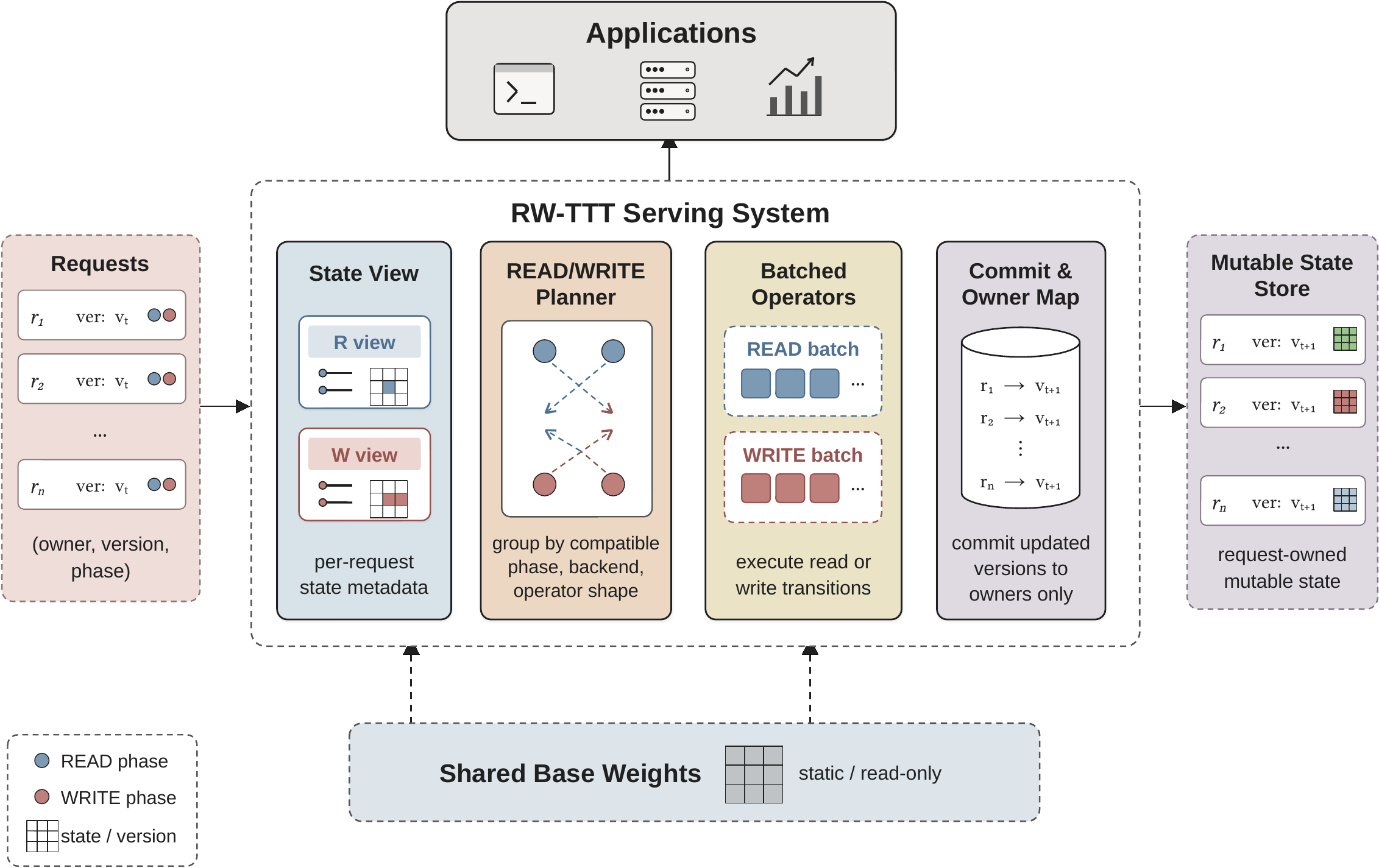}
\caption{\sys{} serving architecture. Requests carry owner-versioned mutable
state alongside ordinary KV cache. The planner automatically groups compatible
READ/WRITE transitions by backend type, operator shape, placement, and valid
owner version; batched operators return outputs and updated state through the
owner map. The figure shows serving semantics, not the backend update rule.}
\label{fig:runtime_stack}
\end{figure*}

\subsection{Serving a Mutable TTT Step}

Figure~\ref{fig:request_flow} shows the per-request transition that the runtime
must preserve. At each decode step, request~$r$ reads its owner-visible TTT
state~$s_r^v$ and evaluates the shared base model with that state. A
\textsc{Read} applies $s_r^v$ and leaves version~$v$ visible to the next step.
A \textsc{Write} consumes backend-specific update evidence~$u_r$ and produces a
dirty candidate state $\tilde{s}_r^{v+1}$, which becomes visible only after
commit. Legal serving must therefore preserve output-to-request mapping,
per-request READ/WRITE order, and owner-local version transitions
(Section~\ref{sec:serving-model}).

Figure~\ref{fig:runtime_stack} places this contract in the serving stack.
Requests carry KV/cache state plus \texttt{TTTState} metadata: owner
id, backend type, version, placement, and backend metadata. Payloads such as
\texttt{FastWeightState}, \texttt{DeltaAdapterState}, and
\texttt{TailBufferState} represent fast weights, adapter deltas, and update
evidence. A \texttt{StateView} exposes the executable slice for the next step:
read views are immutable, while write views carry update inputs and mark the
state dirty until commit.

The planner only observes each request's next transition. It batches compatible
effects and backend shapes automatically, checks that every owner is at its
expected committed version, and uses the owner map~$\mu$ to connect physical
batch slots back to request semantics. For group $G=\{r_b\}_{b=1}^{B_G}$ with
$\mu(b)=r_b$, READ outputs keep all versions unchanged, while WRITE outputs
commit only the owners in $\mu(G)$. Selective commit and rollback use the same
owner-indexed state table by snapshotting $c_r^v \leftarrow s_r^v$ before
speculative writes.

\subsection{Primitives}

Table~\ref{tab:runtime_primitives} lists the serving-visible primitive groups.
A TTT backend owns the update rule; \sys{} fixes how state is applied, when a
candidate version is produced, and how successful writes become owner-visible.

\begin{table}[t]
\centering
\scriptsize
\setlength{\tabcolsep}{2.5pt}
\renewcommand{\arraystretch}{1.08}
\begin{tabularx}{\columnwidth}{@{}l c >{\raggedright\arraybackslash}X >{\raggedright\arraybackslash}X@{}}
\toprule
Primitive & Ph. & State & Contract \\
\midrule
\textsc{ApplyState}
& R
& weights / deltas
& apply; keep version \\
\textsc{TailBufferUpdate}
& R
& tail metadata
& append; no version bump \\
\midrule
\textsc{BoundaryUpdate}
& W
& TTT payload
& compute dirty state \\
\textsc{Commit/Rollback}
& C
& owner/version log
& publish; restore; fork \\
\bottomrule
\end{tabularx}
\caption{Core read-write TTT serving primitives. R/W/C denote read, write, and
control phases. The backend supplies the payload and update rule; the serving
runtime preserves owner/version semantics.}
\label{tab:runtime_primitives}
\end{table}

\paragraph{READ.}
The backend decides how $s_r^v$ affects model computation: fast weights,
low-rank deltas, or another learner state. The runtime requires only that READ
views are immutable and returned with the same version. Tail-buffer maintenance
belongs here because it records request-local evidence for a later update
boundary without publishing a new owner-visible TTT state version.

\paragraph{WRITE.}
The update evidence $u_r$ is backend-specific. \sys{} enforces that the update
reads owner~$r$'s committed version~$v$ and produces one dirty candidate state
for that owner. The candidate is invisible to later READs until committed.
The Triton experiments in Section~\ref{sec:experiments} target WRITE-side
update and selective-commit kernels because ordinary static-weight kernels do
not cover this state-management path.

\paragraph{CONTROL.}
\textsc{Commit/Rollback} controls visibility. Commit publishes a dirty state as
owner~$r$'s next version only after the WRITE group succeeds; rollback restores
the checkpointed owner slot and version; fork creates a separate owner/version
lineage. The version counter changes here, not inside the backend update, and
only owner-indexed TTT records are modified.

\subsection{TTT-Aware Batch Planning}

For each ready transition~$e_i$, the planner computes
$\kappa_i=(\rho_i,\tau_i,\sigma_i,\pi_i)$, covering effect, backend type,
operator shape, and placement. The owner-local version~$v_i$ is checked against
the committed version~$V(r_i)$, but different owners need not have equal numeric
versions. Requests with different keys are never co-issued, so \textsc{Read} and
\textsc{Write} steps are separated by construction. For each key, the planner
emits a group at target batch size~$B$ or waits at most~$w$ decode steps for
compatible arrivals. When the wait budget expires, it issues the current legal
prefix after owner-version and owner-map checks. This changes only which
independent owners are co-issued, not any request's local order or backend
update rule.

\paragraph{Serving loop.}
Algorithm~\ref{alg:main_loop} summarizes the per-iteration control flow.

\begin{algorithm}[t]
\small
\caption{\sys{} TTT-aware serving loop}
\label{alg:main_loop}
\begin{algorithmic}[1]
\Require active requests $\mathcal{R}$, target batch size $B$, wait budget $w$
\While{$\mathcal{R}$ is not empty}
  \State $\mathcal{E}\leftarrow\emptyset$
  \ForAll{$r\in\mathcal{R}$}
    \State $v_r \leftarrow \Call{View}{r.\texttt{TTTState}, r.\texttt{metadata}}$
    \State $e_r \leftarrow \Call{NextStep}{r, v_r}$
    \State $\mathcal{E}\leftarrow \mathcal{E}\cup\{e_r\}$
  \EndFor
  \State $\mathcal{G}\leftarrow \Call{LegalGroups}{\mathcal{E}, B, w}$
  \ForAll{$g\in\mathcal{G}$}
    \State $(Y,\Delta S) \leftarrow \Call{ExecuteOperatorGroup}{g}$
    %\begingroup\color{blue}
    \State $\Call{ReturnOutputs}{Y, g.\texttt{owner\_map}}$
    %\endgroup
    % \begingroup\color{red}
    % \If{execution fails}
    %   \If{$g.\texttt{effect}=\textsc{Write}$}
    %     \State $\Call{Rollback}{g.\texttt{owner\_map}}$
    %   \EndIf
    %   \State \textbf{continue}
    % \EndIf
    % \endgroup
    \If{$g.\texttt{effect}=\textsc{Write}$}
      \State $\Call{CommitVersions}{\Delta S, g.\texttt{owner\_map}}$
    \EndIf
    \State $\Call{UpdateKVAndTailMetadata}{g}$
    % \begingroup\color{red}
    % \State $\Call{ReturnOutputs}{Y, g.\texttt{owner\_map}}$
    % \endgroup
  \EndFor
\EndWhile
\end{algorithmic}
\end{algorithm}

% \rev{WRITE outputs are returned only after commit. Operators return logits $z$, and probabilities
% $p=\operatorname{softmax}(z)$ are introduced only for normalization or sampling.}

\subsection{Backend Instantiations}

This interface lets multiple TTT styles share the same serving contract. A
backend exposes a request-owned payload, immutable read view, write evidence for
its update rule, compatibility metadata (backend type, operator shape, and
placement), and commit metadata for the owner slot and candidate version. The
runtime uses only this interface; the backend retains its learning formula.

For example, a fast-weight backend stores mutable fast weights in
\texttt{FastWeightState}; a LoRA-style backend stores request-owned low-rank
deltas in \texttt{DeltaAdapterState}. Both expose \textsc{Read} application,
\textsc{Write} update, owner/version commit metadata, and compatibility keys.
Other streaming learners fit the same runtime when they expose equivalent
read/write transitions.

\section{Evaluation}
\label{sec:experiments}

We organize the evaluation around five questions: \ding{182} Is the sequential serving collapse caused by mutable-state batching
rather than per-request TTT cost? \ding{183} How much one-GPU serving capacity does \sys{} recover? \ding{184} Does \sys{} preserve the adapted checkpoint's long-context behavior? \ding{185} Do owner or version correctness checks hold under numerical and state-contract tests? \ding{186} How broadly does the same read-write contract apply beyond the headline trace?

\subsection{Experimental Setup and Evidence Scope}

We evaluate a Qwen3-4B In-Place-TTT fast-weight checkpoint
\citep{yang2025qwen3,feng2026inplacettt}. Headline serving results use one GPU,
BF16 cached decode, eight request-private CUDA streams, 4096 prompt tokens,
512 decode tokens, and TTT chunk size 128. Headline throughput is measured
over the decode serving window after prompt state is available. RW-TTT timing
within that window includes grouping, waiting, fallback, and state movement;
prefill, process startup, and model loading are excluded.

\paragraph{Baselines.}
\emph{Sequential In-Place TTT} (29.51 tok/s) is the original
correctness-preserving serial mutable-state execution: one stream at a time.
\emph{Naive sequential replicas} place multiple unmodified batch-1 TTT
processes on one GPU; at three replicas ($\sim$32.50 GiB) this reaches 79.78
tok/s and is the memory-matched deployment reference.
\paragraph{Evidence scope.}
We separate throughput measurement from behavior validation. The throughput harness uses the decode-window scope defined above.
Full task behavior is evaluated on RULER-32K/64K~\citep{hsieh2024ruler},
because throughput measurement and long-context task execution answer different
validation questions. Mutable-state correctness is further validated by
numerical gates and contract stress tests
(\S\ref{sec:correctness_gates}).

\subsection{Batching Bottleneck Diagnosis}
\label{sec:bottleneck}
We first isolate single-request overhead from batching loss. The measured
\sys{} single-request overhead is small, so the sequential serving collapse is
primarily a batching failure.

Controlled batch-1 timing shows that \sys{}'s per-request execution overhead is
modest: median 1.4\% total, 2.1\% prefill, and 0.4\% decode. The serving
bottleneck is therefore the inability to safely batch request-owned mutable
streams. The sequential path yields only 29.51 tok/s for eight concurrent
requests because it serializes all mutable-state effects.

\subsection{GPU Serving Throughput}
\label{sec:throughput}
We next measure aggregate throughput under the one-GPU memory class. \sys{}
recovers most of the lost serving capacity and remains effective under higher
write pressure.

\begin{table}[t]
\centering
\scriptsize
\setlength{\tabcolsep}{2.5pt}
\resizebox{\columnwidth}{!}{%
\begin{tabular}{@{}llrrrr@{}}
\toprule
\multirow{2}{*}{Method}
& \multirow{2}{*}{Trace}
& \multicolumn{1}{c}{Throughput}
& \multicolumn{1}{c}{Peak}
& \multicolumn{2}{c}{Speedup} \\
\cmidrule(lr){5-6}
& & \multicolumn{1}{c}{tok/s}
& \multicolumn{1}{c}{GiB}
& \multicolumn{1}{c}{vs. serial}
& \multicolumn{1}{c}{vs. replicas} \\
\midrule
Serial TTT & uniform & 29.51 & --- & 1.00$\times$ & --- \\
3 serial replicas & uniform & 79.78 & 32.50 & 2.70$\times$ & 1.00$\times$ \\
Phase grouping & uniform & 226.24 & 33.39 & 7.67$\times$ & 2.84$\times$ \\
\sys{} full & uniform & \textbf{274.61} & 33.53 & \textbf{9.31$\times$} & \textbf{3.44$\times$} \\
\sys{} full & bursty update & 177.55 & 33.38 & 6.02$\times$ & 2.23$\times$ \\
\sys{} full & all-update stress & 174.66 & 33.42 & 5.92$\times$ & 2.19$\times$ \\
\bottomrule
\end{tabular}}
\caption{One-GPU serving throughput for request-owned TTT state with eight
streams, 4096 prompt tokens, and 512 decode tokens. Throughput is measured over the decode serving window defined above. The first four rows form the headline comparison against
serial execution and memory-matched replicas; the final two rows stress the
scheduler with higher update pressure.}
\label{tab:serving_main}
\end{table}

Phase-compatible grouping reaches 226.24 tok/s, providing most of the
throughput gain over serial serving. Full RW-TTT reaches 274.61 tok/s on the same workload,
21.4\% higher; the owner-indexed commit ablation reaches 226.25 tok/s
(Appendix~\ref{app:serving_full_rows}), indicating negligible commit overhead.

\begin{figure}[t]
\centering
\includegraphics[width=\columnwidth]{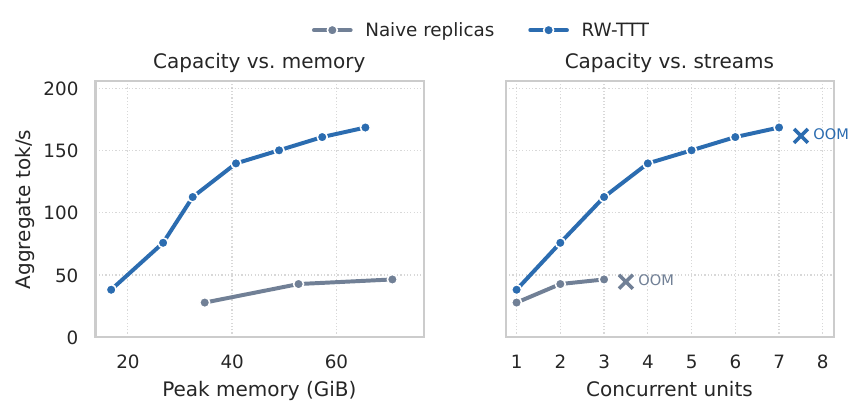}
\caption{Scaling under a 16K prompt.
Left: feasible runs ordered by peak memory. Right: same runs ordered by
concurrent units. \sys{} keeps request-private streams inside one read-write runtime.
Seven streams succeed at 168.66 tok/s and 65.58 GiB; the eighth exceeds
single-GPU memory, so the marked OOM is a context/concurrency capacity frontier.}
\label{fig:serving_scaling_two_axis}
\end{figure}

Table~\ref{tab:serving_main} reports the main throughput and stress
comparisons. The full \sys{} path achieves 274.61 tok/s:
\textbf{9.31$\times$} over the sequential path and \textbf{3.44$\times$} over
memory-matched replicas. The uniform trace contains 4064 read-owner steps and
32 update-owner steps over 4096 generated tokens. Figure~\ref{fig:serving_scaling_two_axis}
shows the scaling under the fixed memory and concurrent setting. Under bursty-update and
all-update stress traces, \sys{} remains above 5.9$\times$ the sequential
baseline, showing that throughput recovery is not limited to favorable
read-dominant schedules.

With one full-model worker and eight streams per rank on a single
eight-H800 node, data-parallel \sys{} scales aggregate uniform throughput from
269.55 tok/s at $\mathrm{DP}=1$ to 1982.69 tok/s at $\mathrm{DP}=8$
(7.36$\times$; 91.9\% scaling efficiency).

\paragraph{Dynamic traffic and bounded waiting.}
A controlled fixed-prefill scheduler simulation retains 95.3\%, 97.0\%, and
79.4\% of uniform throughput under Poisson arrivals, adversarial phase skew,
and heterogeneous decode. On the phase-skew trace, $w=4$ nearly matches $w=8$
throughput while reducing p95 decode-ready-to-first-token latency from 198.65
to 107.81 ms. The six-point wait-budget sweep is in
Appendix~\ref{app:dynamic_wait}.

\subsection{Long-Context Behavior Preservation}
\label{sec:behavior}
We then check whether executing the adapted checkpoint through \sys{} changes
long-context task behavior. We compare against the original In-Place TTT path. 
\begin{table}[t]
\centering
\small
\begin{tabular}{lrr}
\toprule
Benchmark & In-Place TTT & \sys{} \\
\midrule
RULER-32K & 76.30 & 76.41 \\
RULER-64K & 67.72 & 67.72 \\
\bottomrule
\end{tabular}
\caption{Full RULER task scores for the adapted checkpoint executed by the
original In-Place TTT path and by \sys{}. Scores are averages over 13 tasks
with 100 examples each; per-task breakdowns are in
Appendix~\ref{app:ruler_behavior}.}
\label{tab:behavior_main}
\end{table}
Table~\ref{tab:behavior_main} reports full RULER evaluation scores. The same
adapted checkpoint is executed by the original In-Place TTT path and by \sys{}
on RULER, a 13-task long-context evaluation suite~\citep{hsieh2024ruler}. The
32K score is essentially unchanged (76.41 vs.\ 76.30) and the 64K score is
identical (67.72), with the full per-task breakdown retained in
Appendix~\ref{app:ruler_behavior}.

\subsection{Correctness Checks}
\label{sec:correctness_gates}
We combine task scores, calibrated numerical gates, and explicit state-contract
stress tests to check the owner/version contract. All three evidence layers
support the same correctness conclusion, showing in Table~\ref{tab:correctness_gates}.

\begin{table}[t]
\centering
\small
\resizebox{\columnwidth}{!}{%
\begin{tabular}{lll}
\toprule
Check & Scope & Result \\
\midrule
Task behavior & RULER-32K/64K & preserved \\
NLL/PPL & 9 configs $\times$ 512 tokens & 9/9 pass \\
Relative drift & $\alpha{=}1.0$ gate & pass; top5\_min 1.0 \\
State-contract stress & 5 injected failures & 5/5 pass \\
\bottomrule
\end{tabular}}
\caption{Correctness evidence summary. RULER checks task-level behavior; the
NLL/PPL and relative-drift gates check calibrated numerical agreement; the
state-contract row injects owner/version failure modes. Detailed RULER scores
appear in Appendix~\ref{app:ruler_behavior}.}
\label{tab:correctness_gates}
\end{table}

These layers address complementary failure modes. RULER scores validate that
the adapted model produces correct task outputs. The NLL and relative gates
validate numerical fidelity of the batched execution path. State-contract
stress validates owner/version/rollback invariants that average scores may not
expose. Together, these checks target task preservation, calibrated numerical
fidelity, and hard state invariants rather than relying on exact BF16 token
identity as the sole correctness criterion.

\subsection{Operator Evidence}
\label{sec:supporting}
Finally, we look beyond the headline 4K serving trace. The same contract
exposes the 16K memory boundary, carries a second adapter-shaped mutable state,
and identifies WRITE-side operator targets.

\begin{figure}[t]
\centering
\includegraphics[width=0.92\columnwidth]{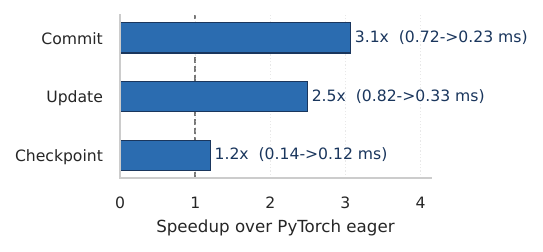}
\caption{WRITE-side operator speedups from Triton kernels over PyTorch eager.
Parentheses show latency before and after kernelization in milliseconds.}
\label{fig:operator_microbench}
\end{figure}

Figure~\ref{fig:operator_microbench} provides local evidence for the read-write
operator surface induced by mutable TTT state. Triton kernels reduce WRITE-side
latency for selective commit, dense update, and checkpoint write by
\(3.1\times\), \(2.5\times\), and \(1.2\times\), respectively. These
microbenchmarks are not the end-to-end serving denominator; they show that the
unusual optimization surface lies in owner-indexed commit, state update, and
checkpointing rather than ordinary shared-weight projection.

\section{Discussion}

\paragraph{Phase skew remains hard.}
Requests are not fixed as boundary or non-boundary streams; each stream alternates
between READ and WRITE steps. When updates align, the planner forms dense WRITE
batches, but phase skew fragments groups. The bounded wait budget trades latency
for group density without changing semantics, leaving skew-aware scheduling as
future work.

\paragraph{State movement and layout.}
Read-write TTT shifts the serving surface from shared-weight projection to
mutable-state commit, checkpoint, and movement. State packing and ownership
tracking grow with the number of owners, making state layout and memory
management the next bottleneck after batching is restored. Execution becomes bandwidth-bound when state-movement time exceeds the
compute time saved by batching.

\paragraph{Operator kernels.}
Microbenchmarks show wins for selective commit and dense update operators,
while read-path projection does not beat \texttt{torch.bmm}. Thus WRITE-side
kernelization is a follow-up, not a substitute for the serving contract.

\paragraph{Validation scope.}
Full validation uses fast-weight In-Place-TTT checkpoints with a
sequential reference and long-context RULER tasks. LoRA-style experiment
tests the same owner/version contract on adapter-shaped payloads, but is not a
full downstream adapter-serving evaluation.

\section{Related Work}

\paragraph{KV/prefix-cache serving.}
Modern LLM serving has optimized the shared-weight contract through
iteration-level scheduling \citep{yu2022orca}, paged KV-cache management
\citep{kwon2023vllm}, chunked-prefill/decode overlap
\citep{agrawal2023sarathi,agrawal2024sarathiserve}, prefill/decode
disaggregation \citep{zhong2024distserve}, program-level scheduling and cache
reuse \citep{zheng2024sglang}, and further refinements in high-throughput
generation, co-serving, and elastic disaggregation
\citep{holmes2024deepspeedfastgen,qiao2024conserve,ruan2025dynaserve,wan2025bros}.
These systems manage dynamic \emph{activation}-side state while model-side parameters remain shared and read-only. \sys{} keeps
the scheduling problem but changes the object being scheduled: each request
also owns model-side state whose version may change during generation.

\paragraph{Adapter serving.}
S-LoRA and Punica make heterogeneous request-specific model state a
first-class serving concern \citep{hu2022lora,sheng2023slora,chen2023punica,huggingfacepeft}.
Their core contract is \emph{read-side heterogeneity}: requests apply
different low-rank weights, while the serving stack manages placement and
batched execution, but adapters remain read-only during generation. Read-write
TTT builds on the same insight that small request-specific state can be
served efficiently, then adds online mutation: a LoRA-style TTT state can be
read like an adapter, dirtied by an update event, committed under a new
version, and read again by the same request. Our LoRA-style stress test confirms
that the same interface carries adapter-shaped payloads.

\paragraph{Test-time training.}
Test-time training adapts models using information available at inference time
\citep{sun2020testtime}. Recent work uses test-time learning inside
recurrent-style hidden states \citep{sun2024tttlayers}, while In-Place-TTT
applies online updates to selected LLM layers during long-context generation
\citep{feng2026inplacettt}. These are backend choices for producing mutable
state and update rules. \sys{} targets the missing serving-system layer needed
when such methods read and write request-owned state during generation.

\section{Conclusion}

Test-time training introduces request-owned mutable model state that breaks
the static-weight batching contract behind current LLM serving systems.
\sys{} restores batching through a read-write serving contract: each decode
step is tagged with its owner and READ/WRITE effect, only compatible phases
are grouped, and updated state versions are committed exclusively to owning
requests. On eight concurrent streams with a fast-weight In-Place-TTT
checkpoint, this recovers 274.61 aggregate tok/s---9.31$\times$ over the
sequential TTT path and 3.44$\times$ over memory-matched independent
replicas---while preserving adapted RULER-32K/64K behavior and passing
owner/version correctness gates. The contribution is a serving contract and
batching runtime for request-owned mutable state, not a new TTT learning rule
or kernel system; \sys{} separates TTT backend design from the serving
semantics needed to batch online-adaptive model state.

\section*{Limitations}
RW-TTT evaluates serving for existing online TTT behavior rather than proposing a new adaptation rule. We hold the backend update rule fixed and ask whether request-owned mutable state can be batched without changing adapted behavior. Full task evaluation uses fast-weight In-Place-TTT checkpoints, which provide a sequential reference and long-context evaluation; the LoRA-style experiment tests the same owner/version READ/WRITE contract on adapter-shaped mutable state. Throughput is reported for the prototype’s one-GPU memory class and eight-stream workload, compared against sequential TTT and memory-matched unmodified replicas. Multi-node, tensor-parallel, and pipeline-parallel execution remain untested. Static-weight serving follows a different contract: RW-TTT targets request state that is read and written during generation.

\section*{Acknowledgments}
This work is funded in part bythe HKUST Start-up Fund (R9911), Theme-based Research Scheme grant (T45-205/21-N), the InnoHK initiative of the Innovation andTechnology Commission of the Hong KongSpecial Administrative Region Government,and the research funding under HKUST-DXMAl for Finance Joint Laboratory (DXM25EG01).

\bibliography{references}

\clearpage
\appendix
\section{Full Serving Rows}
\label{app:serving_full_rows}

\begin{table}[H]
\centering
\scriptsize
\setlength{\tabcolsep}{2.5pt}
\renewcommand{\arraystretch}{1.08}
\begin{tabularx}{\columnwidth}{@{}>{\RaggedRight\arraybackslash}p{0.35\columnwidth}>{\RaggedRight\arraybackslash}Xrr@{}}
\toprule
Method / unit & Grouping mode & tok/s & Speedup \\
\midrule
Serial TTT / 8 streams & serial groups & 29.51 & 1.00 \\
3 serial replicas / 3 replicas & independent batch-1 & 79.78 & 2.70 \\
\sys{} phase grouping / 8 streams & phase-compatible & 226.24 & 7.67 \\
\sys{} + owner commit / 8 streams & phase-compatible & 226.25 & 7.67 \\
\sys{} full read/write / 8 streams & read/write groups & \bfseries 274.61 & \bfseries 9.31 \\
\sys{} bursty update / 8 streams & phase-skewed & 177.55 & 6.02 \\
\sys{} all-update stress / 8 streams & dense write & 174.66 & 5.92 \\
\bottomrule
\end{tabularx}
\caption{Full-model prototype serving-harness results. Speedup is relative
to the serial In-Place TTT path on one GPU.}
\label{tab:serving_harness_full}
\end{table}

Table~\ref{tab:serving_harness_full} serving harness shows that phase-compatible batching is the dominant
source of throughput recovery: \sys{} phase grouping improves aggregate
throughput from 29.51 to 226.24 tok/s, while adding owner-commit semantics
does not introduce measurable overhead. The full read/write system reaches
274.61 tok/s, or 9.31$\times$ over serial execution, and remains above
5.9$\times$ speedup even under bursty-update and all-update stress settings.

\section{Wait-Budget Sensitivity}
\label{app:dynamic_wait}

We sweep the bounded wait budget $w$ on the phase-skew trace using eight
streams, 4096-token prompts, nominal 512-token decoding, and 128-token TTT
boundaries. The controlled scheduler simulation uses one service-time
calibration across all rows and averages results over five seeds.

\begin{table}[H]
% \color{red}
\centering
\scriptsize
\setlength{\tabcolsep}{1.6pt}
\begin{tabular}{@{}rrrrr@{}}
\toprule
$w$ & Sim. tok/s & First-token p95 (ms) & Completion p95 (s) & Violations/starvation \\
\midrule
0 & 272.57 & 50.01 & 15.03 & 0 \\
1 & 275.14 & 67.73 & 14.89 & 0 \\
2 & 275.14 & 86.99 & 14.89 & 0 \\
4 & 276.44 & 107.81 & 14.82 & 0 \\
8 & 276.59 & 198.65 & 14.81 & 0 \\
16 & 274.61 & 198.65 & 14.92 & 0 \\
\bottomrule
\end{tabular}
\caption{Wait-budget sweep on the simulated phase-skew trace. Latency is
measured from decode-ready to the first generated token; the final column
aggregates bounded-wait, owner/version, and starvation checks.}
\end{table}

\section{RULER Per-Task Scores}
\label{app:ruler_behavior}

\begin{table}[H]
\centering
\small
\resizebox{\columnwidth}{!}{%
\begin{tabular}{lrrr}
\toprule
Task & In-Place TTT & Repeat & \sys{} \\
\midrule
ruler\_qa\_squad\_32k & 45.00 & 45.00 & 46.00 \\
ruler\_qa\_hotpotqa\_32k & 48.00 & 48.00 & 48.00 \\
ruler\_niah\_single\_1\_32k & 100.00 & 100.00 & 100.00 \\
ruler\_niah\_single\_2\_32k & 89.00 & 89.00 & 90.00 \\
ruler\_niah\_single\_3\_32k & 99.00 & 99.00 & 99.00 \\
ruler\_niah\_multikey\_1\_32k & 66.00 & 66.00 & 66.00 \\
ruler\_niah\_multikey\_2\_32k & 82.00 & 82.00 & 82.00 \\
ruler\_niah\_multikey\_3\_32k & 32.00 & 32.00 & 32.00 \\
ruler\_niah\_multivalue\_32k & 97.75 & 97.75 & 97.75 \\
ruler\_niah\_multiquery\_32k & 94.50 & 94.50 & 94.50 \\
ruler\_vt\_32k & 99.80 & 99.80 & 99.60 \\
ruler\_fwe\_32k & 96.00 & 96.00 & 95.67 \\
ruler\_cwe\_32k & 42.80 & 42.80 & 42.80 \\
\bottomrule
\end{tabular}}
\caption{Full step-aligned 32K RULER task scores for the adapted fast-weight
checkpoint.}
\end{table}

\begin{table}[H]
\centering
\small
\resizebox{\columnwidth}{!}{%
\begin{tabular}{lrrr}
\toprule
Task & In-Place TTT & \sys{} & Delta \\
\midrule
ruler\_qa\_squad\_64k & 44.00 & 44.00 & 0.00 \\
ruler\_qa\_hotpotqa\_64k & 44.00 & 44.00 & 0.00 \\
ruler\_niah\_single\_1\_64k & 100.00 & 100.00 & 0.00 \\
ruler\_niah\_single\_2\_64k & 59.00 & 59.00 & 0.00 \\
ruler\_niah\_single\_3\_64k & 96.00 & 96.00 & 0.00 \\
ruler\_niah\_multikey\_1\_64k & 34.00 & 34.00 & 0.00 \\
ruler\_niah\_multikey\_2\_64k & 85.00 & 85.00 & 0.00 \\
ruler\_niah\_multikey\_3\_64k & 22.00 & 22.00 & 0.00 \\
ruler\_niah\_multivalue\_64k & 95.25 & 95.25 & 0.00 \\
ruler\_niah\_multiquery\_64k & 94.75 & 94.75 & 0.00 \\
ruler\_vt\_64k & 100.00 & 100.00 & 0.00 \\
ruler\_fwe\_64k & 94.00 & 94.00 & 0.00 \\
ruler\_cwe\_64k & 12.40 & 12.40 & 0.00 \\
\bottomrule
\end{tabular}}
\caption{Full 64K RULER task scores for the matching
\texttt{qwen3-4b-64k-ttt-step1250} checkpoint.}
\end{table}

\section{PEFT/LoRA Adapter-State Path Check}
\label{app:peft_path}

The LoRA-style path maps rank-8 PEFT adapter state to
\texttt{DeltaAdapterState} and exercises the same ownership, versioning,
READ/WRITE, and commit operations. Four materialized adapter/domain pairs pass
exact apply/update equivalence checks, with NLL changes below
$3.4{\times}10^{-3}$. This checks the contract on adapter-shaped mutable state;
it is not a downstream LoRA-TTT quality benchmark.

\end{document}